\documentclass[a4paper]{article}

\usepackage{INTERSPEECH2022}
\usepackage[super]{nth}
\usepackage{amsmath, nccmath}
\usepackage{multirow}
\usepackage{array,ragged2e}
\newcolumntype{P}[1]{>{\RaggedRight\arraybackslash}p{#1}}
\usepackage{booktabs}
\newcommand{\ra}[1]{\renewcommand{\arraystretch}{#1}}
\usepackage{float}
\restylefloat{table}
\usepackage{tabu, makecell, graphicx, booktabs}

\title{Multilingual Transformer Language Model for Speech Recognition in Low-resource Languages}
\name{Li Miao, Jian Wu, Piyush Behre, Shuangyu Chang, Sarangarajan Parthasarathy}
\address{Microsoft, USA}
\email{\{limia|jianwu|piyush|shchang|sarangp\}@microsoft.com}

\begin{document}

\maketitle
\begin{abstract}

It is challenging to train and deploy Transformer LMs for hybrid speech recognition \nth{2} pass re-ranking in low-resource languages due to (1) data scarcity in low-resource languages, (2) expensive computing costs for training and refreshing 100+ monolingual models, and (3) hosting inefficiency considering sparse traffic. 
In this study, we present a new way to group multiple low-resource locales together and optimize the performance of Multilingual Transformer LMs in ASR. Our Locale-group Multilingual Transformer LMs outperform traditional multilingual LMs along with reducing maintenance costs and operating expenses. Further, for low-resource but high-traffic locales where deploying monolingual models is feasible, we show that fine-tuning our locale-group multilingual LMs produces better monolingual LM candidates than baseline monolingual LMs.

\end{abstract}

\section{Introduction}

Automatic Speech Recognition (ASR) usually involves two passes. The first-pass acoustic models and n-gram language models generate n-best hypotheses from the global search space \cite{yu2014automatic}. In the second pass, Neural Network Language Models (NNLM) are widely used to re-rank the n-best hypotheses \cite{erdogan2016multi-channel}. It has been demonstrated that re-ranking using NNLM is effective at reducing WER (Word Error Rate) \cite{6854535}, with Transformer language models producing state-of-the-art results in re-ranking \cite{DBLP:journals/corr/abs-1905-04226}.

Today our ASR system supports 100+ locales, but re-ranking is only applied to a few high-resource locales, even though we have proven the higher benefits of re-ranking for low-resource locales like Slovenian. The key challenges today are: (1) the low-resource locales’ training data is scarce, which limits our capacity to train the NNLM, (2) it is computationally expensive to train and regularly refresh 100+ monolingual re-ranking models, one for each locale, (3) it is prohibitively expensive and inefficient to host these monolingual models in production, as traffic can be sparse, but each model ends up taking memory and compute to host across all our Speech clusters.

Multilingual Transformer language models \cite{DBLP:journals/corr/abs-1901-07291} \cite{DBLP:journals/corr/abs-1911-02116} \cite{DBLP:journals/corr/abs-1906-01502} provide a very effective general solution to support ASR with pre-trained components and data sources that can be shared across multiple languages. When applied blindly, however, multilingual transformer language models may not always match or beat the monolingual models. We found that grouping multiple akin locales can optimize performance, especially when dealing with low-resource locales. As a result, our Locale-group LMs outperform the general multilingual solutions.

This key insight has helped us tackle the above-listed challenges. Locales with limited resources (scarce data) can benefit from all the data available for their locale group. We would only need to train and maintain a few locale-group Transformer LMs and still can attain locale coverage of 100+ locales for re-ranking, and fewer overall \nth{2} pass LMs result in hosting and scaling efficiencies across our clusters in production.

In addition, our key finding regarding grouping low-resource locales has been found to work in other related domains, such as improving capitalization and punctuation in recognition outputs, with potential future applications beyond speech.

\begin{figure*}[t]
  \centering
  \includegraphics[width=\linewidth]{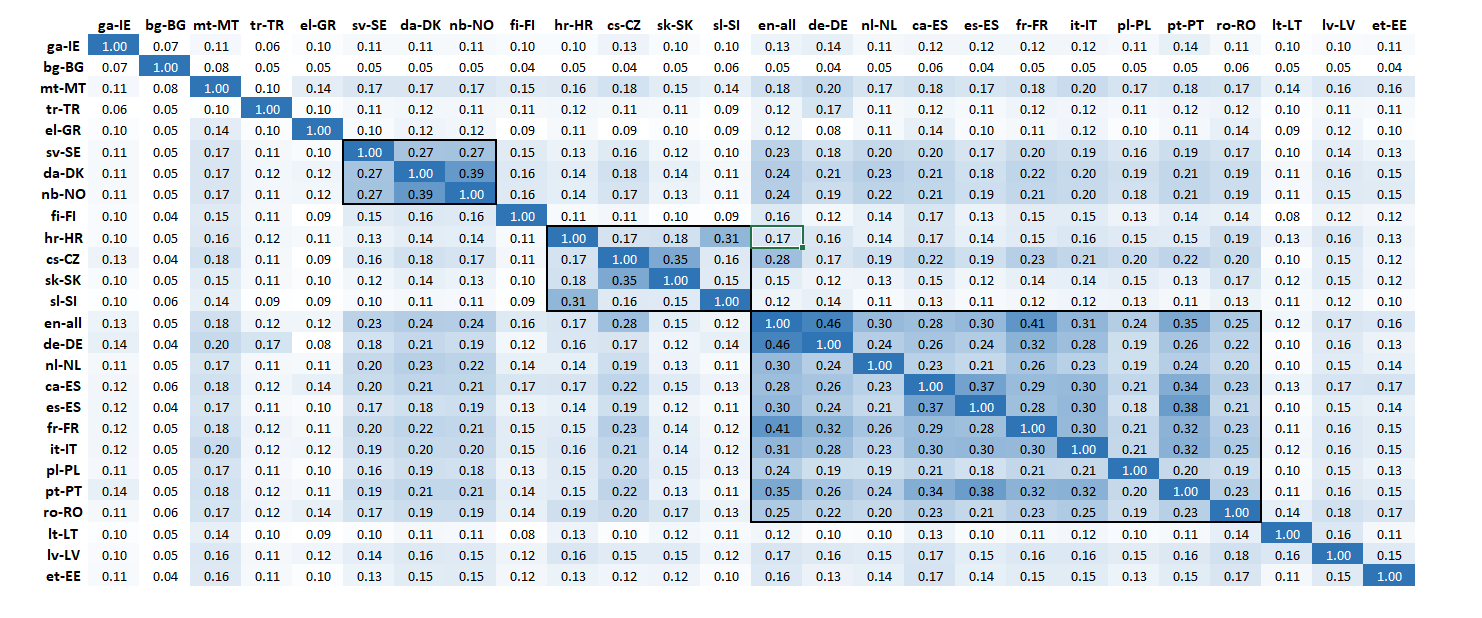}
  \caption{Example of lexical similarity scores across 26 European languages.}
  \label{fig:language_similarity}
\end{figure*}

\section{Related Work}
\textbf{Multilingual/Cross-lingual.} 
The effectiveness of sentence encoders' generative pre-training was first demonstrated for English natural language processing \cite{radford2018improving} \cite{DBLP:journals/corr/abs-1810-04805} \cite{DBLP:journals/corr/abs-1801-06146}. Multiple approaches have since been proposed to extend it to multilingual/cross-lingual pretraining and show the success in transfer learning, such as mBERT \cite{DBLP:journals/corr/abs-1810-04805}, XLM \cite{DBLP:journals/corr/abs-1901-07291}, XLM-R \cite{DBLP:journals/corr/abs-1911-02116}, Unicoder \cite{huang2019unicoder}, etc. Large amounts of unlabeled data from multiple languages are used to train these models, with the goal that low-resource languages can benefit from high-resource languages from shared vocabularies and underlying linguistic similarities. mBert trains a BERT model using Wikipedia corpora in 104 languages. XLM introduced a translation language model (TLM) in addition to masked language model (MLM), in which bilingual sentences are concatenated as inputs. To further improve the performance, Unicoder presents three new cross-lingual pre-training tasks, including cross-lingual word recovery, cross-lingual paraphrase classification and cross-lingual masked language model. XLM-R trains exclusively with MLM objective on a huge multilingual dataset at an enormous scale. 

Multilingual is also explored in ASR, primarily from an acoustic perspective. In \cite{pratap2020massively}, a massive multilingual acoustic model trained with more than 50 languages and more than 16,000 hours of audio is proven to improve recognition performance especially in low-resource languages. XLSR \cite{DBLP:journals/corr/abs-2006-13979}, a cross-lingual speech representation learning method is proposed by pre-training a single model from 56,000 hours raw waveform of speech in 53 languages. 

Our work mainly focuses on Multilingual Language Model in ASR \nth{2} pass re-ranking, where a language model score is interpolated with \nth{1} pass LM and AM to select the 1-best recognition candidate. 

\section{Locale-group Transformer LM}

Our proposed approach involves two steps. The first is to identify the underlying language group of the low-resource locale using our data-driven method, and the second is to process the low-resource locales’ data with shareable Byte Pair Encoding (BPE) tokens \cite{DBLP:journals/corr/SennrichHB15} and train the large-scale Locale-group Transformer Language Model.  Whenever we lack enough resources and hardware to support individual model development and deployment, we can choose to deploy the group-based multilingual Transformer Language Model, which provides significant Speech Recognition accuracy improvement, maintenance and cost reduction. 

\subsection{Language Group Identification} \label{ssec:lan_gp_id}

As one of three organizations selected to potentially partner with the European Parliament in 2020, Microsoft developed a real-time AI-based tool for live transcription and translation of debates. To identify the underlying language groups for 26 European languages, we proposed a two-step data-driven mechanism.

Firstly, we computed bi-lingual similarity score, which can be a measure of the number of phonemes, words, phrases, or the like that are present in both locales. The Figure \ref{fig:language_similarity} shows an example of bi-lingual lexical similarity scores for 26 European languages, where higher score indicates closer linguistic relations between the languages. We observed code-switching/loanwords to be common, especially in English. In contrast, per our experiments Bulgarian (bg-BG) does not appear to be close to any other languages based on our collected data. We suspect that some of the data skew was caused by source data filtering.

Secondly, we applied vector-based clustering techniques to categorize similar languages together based on similarity score vectors. As shown in Table \ref{tab:lan_gp}, this mechanism successfully identifies language family like Balto-Slavic, the group 2, which contains Slovenian, Croatian, Slovak and Czech. Group 3 consists of most high-resource Germanic languages such as English and German, and Latin (Romance) languages like Italian, Spanish and French. In Group 4, Greek is supposed to have its own alphabet - the similarity with other languages is mainly due to code-switching/loanwords. 

\begin{table}[th]
  \ra{1.3}
  \caption{Language groups of 26 European locales}
  \label{tab:lan_gp}
  \centering
  \begin{tabular}{@{}p{1cm}p{5.3cm}@{}}
    \toprule
    \textbf{Group}   & \textbf{Languages} \\
    \midrule
    1 & nb-NO, sv-SE, fi-FI, da-DK \\
    \hline
    2 & sl-SI, hr-HR, cs-CZ, sk-SK \\
    \hline
    3 & en-all, es-ES, nl-NL, fr-FR, ro-RO, ca-ES, it-IT, pt-PT, pl-PL, de-DE \\
    \hline
    4 & bg-BG, lv-LV, lt-LT, ga-IE, et-EE, el-GR, mt-MT, tr-TR \\
    \bottomrule
  \end{tabular}
\end{table}


\subsection{Shareable BPE Tokens}

In our proposed approach, we process all languages with the same shared vocabulary created through Byte Pair Encoding (BPE) \cite{DBLP:journals/corr/SennrichHB15}. We provides several BPE format examples in Table \ref{tab:bpe_example}. This approach can greatly improve the token coverage with limited token set size and standardize the sub-word units across languages that share the same alphabet. For example, with 250K BPE tokens, we achieve almost a 100\% coverage of 350M unique words across 26 languages.

\begin{table}[th]
  \ra{1.2}
  \caption{Examples of text in BPE format}
  \label{tab:bpe_example}
  \centering
  \begin{tabular}{@{}P{1cm}P{2.4cm}P{2.6cm}@{}}
    \toprule
    \textbf{Locale}   & \textbf{Word-based Sent}                         & \textbf{BPE-based Sent}  \\
    \midrule
    English           & ask consternation to my word list                    & ask conster@@ nation to my word list   \\
    \hline
    Irish             & a naoi scoil nach bhfuil seomra acmhainne acu        & a naoi scoil nach bhfuil seomra acmha@@ inne acu   \\
    \hline
    Estonia           & asukoha nimed on sofia ja bulgaaria                  & asukoha ni@@ med on sofia ja bulgaaria   \\
    \bottomrule
  \end{tabular}
\end{table}

\subsection{Transformer Language Model}
\subsubsection{Data Balance}

We compile a language group training dataset with a balancing mechanism and train the Locale-group Multilingual Transformer Language Model. To ensure balanced data coverage for multiple regions within the same family, we sample sentences with multinomially distributed probabilities $\{q\}_{i=1...N}$, similar as how sentences are sampled in \cite{DBLP:journals/corr/abs-1901-07291}, 

\begin{equation}
q_{i} = \frac{p_{i}^\alpha}{\sum_{j=1}^{N}p_{j}^\alpha}  \quad \text{ with } \quad  p_{i} = \frac{n_{i}}{\sum_{k=1}^{N}n_{k}}
\end{equation}

\subsubsection{Locale-group Model Training}
Using balanced language family data, we train the Locale-group Multi-lingual Transformer Language Model, which is similar to the structure used in \cite{Radford2018ImprovingLU}. 

During training, we record the valid loss and perplexity of individual locales, and of the language family. According to our findings, the average loss minimum is within the range of the individual locale's loss minimum, which indicates that the Locale-group Multilingual Transformer Language Model converges for all locales within the identified language group.

\subsubsection{General \& Masked Fine-tuning} \label{sssec:ftmft}

In natural language processing (NLP) and machine translation (MT), fine-tuning a pre-trained language model has become the de facto standard for transfer learning \cite{DBLP:journals/corr/abs-1801-06146} \cite{DBLP:journals/corr/abs-2008-00401}. In our work, our ability to serve efficiently and reduce computation costs will be compromised if we fine-tune the multilingual model towards a target language. However, the possibility of fine-tuning is also worth exploring, as (1) knowledge transfer will be verified if we can achieve better performance with fine-tuned monolingual model than the one trained from scratch, along with a unified training recipe, and (2) certain high-traffic locales will separate themselves from low-resource groups as more data is collected and more traffic is received over time, which allows us to support monolingual model deployment with adequate resources. 

In general fine-tuning, we reserve the pre-trained multilingual model parameters, feed the target locale's data in, and train for several more epochs until it converges to a new minimum.

\begin{figure}[t]
  \centering
  \includegraphics[width=\linewidth]{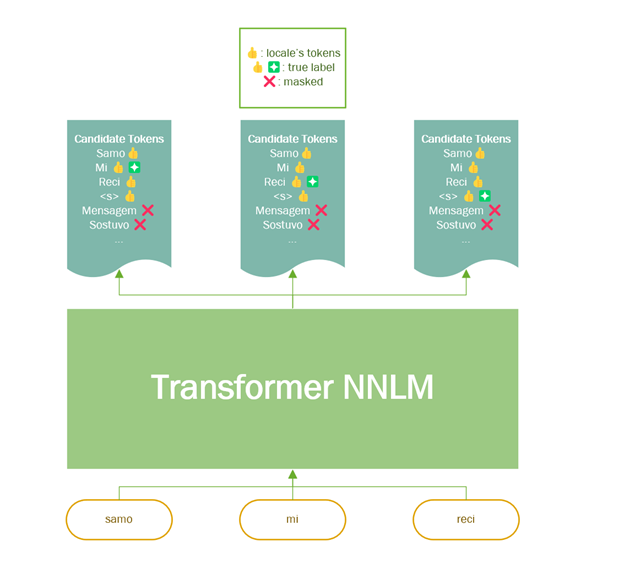}
  \caption{Mask Fine-tuning Diagram with a Croatian example. “Samo mi reci” means “Just tell me” in English.}
  \label{fig:masked_finetune}
\end{figure}

Masked Fine-tuning is designed to force the model to tune to a target locale. For BPE tokens that do not exist in the target locale, we freeze the token embedding updates and set the prediction score to a very small negative number, so the token loss will be close to 0, and avoid paying attention to irrelevant tokens. This process is illustrated in Figure \ref{fig:masked_finetune}. 

\begin{table*}[ht!]
  \ra{1.3}
  \caption{WERR of several low-resource locales}
  \label{tab:werr_table}
  \centering
  \begin{tabular}{@{}P{1.5cm}P{2cm}P{1cm}P{1cm}P{1cm}P{1cm}P{1cm}P{1cm}P{1cm}@{}}
    \toprule
                                 &                    & hr-HR & sl-SI & sk-SK & lt-LT & lv-LV & ro-RO & Avg \\
    \midrule
    \multirow{2}{2em}{Mono}      & LSTM               & 12.15 & 12.13 & 13.21 & 10.12 & 11.78 & 9.28  & 11.45 \\
                                 & Trans              & 16.66 & 15.13 & 16.12 & 14.37 & 15.06 & 11.28 & 14.79 \\
    \hline
    \multirow{2}{2em}{Multi}     & Trans-All          & 13.83 & 13.06 & 15.08 & 10.74 & 11.92 & 9.68  & 12.38 \\
                                 & Trans-Group        & 16.81 & 15.81 & 17.14 & 13.21 & 14.08 & 10.35 & 14.57 \\
    \hline
    \multirow{2}{4em}{Multi+FT}  & Trans-All          & 17.38 & 16.01 & 17.22 & 14.63 & 15.07 & 11.73 & 15.36 \\
                                 & Trans-Group        & 17.6  & 16.21 & 17.42 & 15.21 & 14.97 & 11.48 & 15.50 \\
    \hline
    \multirow{1}{4.3em}{Multi+MFT} & Trans-Group      & 18.22 & 17.83 & 17.75 & 15.15 & 16.13 & 11.48 & 16.09 \\
    \bottomrule
  \end{tabular}
\end{table*}

\section{Experiment}
\subsection{Dataset}
\subsubsection{Train}
We sampled 5B sentences across 26 European locales from our in-house text data corpus, with an average sentence length of 12. All text data are pre-processed into lexical format with our own text-normalization pipeline. Low-resource locales are up-sampled per Section 2.3.1. The same data is used to train the shareable 250K BPE tokens.
\subsubsection{Test}
Test audio data used for word-error-rate reductions (WERR) measurement of each locale consists primarily of dictations and spontaneous conversations. Minimum coverage per locale is 10K sentences. 

\subsection{Model}
We experiment with configurations described in Table \ref{tab:model_config}. 

\begin{table}[th]
  \ra{1.3}
  \caption{Model Configs}
  \label{tab:model_config}
  \centering
    \begin{tabular}{@{}P{3.3cm}P{4cm}@{}}
    \toprule
    \textbf{Config}   & \textbf{Description} \\
    \midrule
    Mono:LSTM               & baseline monolingual Long short-term memory (LSTM) LM \\
    \hline
    Mono:Trans              & monolingual Transformer LM \\
     \hline
    Multi:Trans-All         & all-languages-together Multilingual Transformer LM \\
     \hline
    Multi:Trans-Group       & Locale-group Multilingual Transformer LM \\
     \hline
    Multi+FT:Trans-All      & fine-tuned all-languages-together Multilingual Transformer LM \\
     \hline
    Multi+FT:Trans-Group    & fine-tuned Locale-group Multilingual Transformer LM \\
     \hline
    Multi+MFT:Trans-Group   & masked fine-tuned Locale-group Multilingual Transformer LM \\
    \bottomrule
  \end{tabular}
\end{table}

We train a baseline 1x1024:512 LSTM LM for each language, where 1 is the number of layers, 1024 is the dimensionality of the LSTM state, 512 is the dimensionality of the embedding and also the output dimensionality of the projection layer.

In addition, we trained one multilingual all-languages-together, four multilingual locale-group, and 26 monolingual Transformer language models with the same shareable 250K BPE token set, same data distribution and similar model configuration. These Transformer language models consist of 12 transformer layers, where each transformer layer contains 4096 feedforward dimensions with 16 heads. The warm-up is set to increase the learning rate gradually to improve the convergence of Transformer LMs \cite{DBLP:journals/corr/abs-1804-00247}.

We also applied general fine-tuning and masked fine-tuning as described in section \ref{sssec:ftmft}.

\subsection{Results}

In this work, we mainly report word-error-rate reductions (WERR) on several low-resource locales: Croatian (hr-HR), Slovenian (sl-SI), Slovak (sk-SK), Lithuanian (lt-LT), Latvian (lv-LV) and Romanian (ro-RO). Language groups are described in section \ref{ssec:lan_gp_id}. 

As shown in Table \ref{tab:werr_table}, we observe 3.34\% average WERR improvement because of architecture upgrade from LSTM to Transformer, and more parameters. However, it is challenging to deploy those monolingual Transformer LMs as we discussed three limitations in the beginning.

On the other hand, the Locale-group Transformer LM provides a good solution considering the deployment restrictions. Firstly, with one Transformer LM to serve multiple locales in the same group, we can achieve 3.12\% average WERR gain compared with the LSTM baseline. Secondly, the locale-group model in general outperforms the all-data-together model by 2.19\% when the implicit language similarity information is included, and limited model capacity can spare more attention to the learning of underlying linguistic patterns in similar languages instead of being distracted by irrelevant signals. 

Even more, if we can allocate enough resources to train and deploy dedicated Transformer LM for low-resource locales, we can achieve more with masked fine-tuning. Compared with the monolingual LSTM baseline, our multilingual Locale-group model with masked fine-tuning can provide an additional 4.64\% average WERR. 

\section{Discussion}
\subsection{BPE Token Size}
We also trained models with 64k shareable BPE tokens to evaluate the impact of BPE token size. WERR on individual locales varies, but compared with 250K, the 64K BPE based Locale-group models generally regress around 2.48\%. When we do masked fine-tuning, the gap is reduced to 0.66\%. Our hypothesis is that the average text sequence length are elongated by smaller BPE token size, therefore brings in challenges to learn generic sequence patterns under the same number of model parameters. 

\subsection{Hosting Efficiency}
One key motivation for exploring this idea was to improve production efficiency. Our speech service needs to be deployed globally to tens of clusters. Monolingual and general multi-lingual are two extremes when it comes to hosting costs. Deploying over 100 monolingual models everywhere is quite resource heavy and often wasteful as traffic can be sparse for some locales. On the other hand, general multi-lingual models oversimplify this, at the cost of WERR regression for some locales. Natural data imbalance is also harder to tackle at this scale, and often multi-lingual models can beat monolingual only by increasing model complexity (like adding more Transformer layers). This is also undesirable as this affects individual request latency at serving time. Our approach of locale-group multi-lingual models finds a more optimal point, needing us to deploy only a few locale-group models across our speech clusters, and better WERR than both monolingual and general multi-lingual models in most cases. We keep the option to deploy monolingual or fine-tuned monolingual models for certain high-traffic locales.

\subsection{Parameter Tuning}
We haven't tuned dropout and hyper parameters extensively for all the models. We believe we will achieve more gains with parameter tuning, and plan to explore this in the future work.

\subsection{Application in other areas}
In our hybrid ASR system, there are many other areas where we use monolingual models or general multi-lingual models. We explored the applicability of this locale-grouping technique to Capitalization models. We achieved similar results with locale-group capitalization models outperforming monolingual and general multi-lingual capitalization models.

\section{Conclusion}
Neural Network Language Model (NNLM) is an essential module in hybrid ASR to deliver the optimal recognition accuracy. In this work, we proposed a general and scalable approach to train and deploy large-scale Locale-group Transformer NNLM to support ASR in low-resource languages, where we observed significant accuracy improvement and reduction in model development and maintenance. Further, our fine-tuning experiments show that this locale-grouping helps create better monolingual models for low-resource languages. 

\bibliographystyle{IEEEtran}
\newpage
\bibliography{mybib}

\end{document}